\documentclass[sigconf,nonacm]{acmart}

\usepackage{tikz}
\usepackage{pgfplots}
\pgfplotsset{compat=1.15}
\usetikzlibrary{arrows.meta,positioning,fit,backgrounds,calc}
\usepackage{booktabs}
\usepackage{tabularx}
\usepackage{array}
\usepackage{enumitem}

\AtBeginDocument{%
  \providecommand\BibTeX{{\normalfont B\kern-0.5em{\scshape i\kern-0.25em b}\kern-0.8em\TeX}}}
\begin{document}

\title{Auditing Self-Evolution in Financial Agents:\\Capability Gains, Security Drift, and Execution-Interface Mismatch}

\makeatletter
\def\@ACM@checkaffil{}
\makeatother

\author{Jialong Li}
\authornote{First author.}
\orcid{0009-0005-8347-3958}
\affiliation{%
  \institution{Independent Researcher}
}

\author{Jialing Zhu}
\orcid{0009-0001-3799-8578}
\affiliation{%
  \institution{Independent Researcher}
}

\hypersetup{pdfauthor={Jialong Li, Jialing Zhu}}

\begin{abstract}
Self-evolving agents turn experience into reusable skills, workflows, or memories, but
post-evolution accuracy alone does not show whether learned behaviour preserves previously
correct behaviour or security. We audit SkillOpt, Agent Workflow Memory (AWM), and
ReasoningBank in simulated e-banking using matched benign acquisition trajectories,
sealed evaluation endpoints, execution-grounded checks, and independent state replay.
On Qwen 3.7 Flash, SkillOpt raises benign utility from 0.741 to 0.837, with 23
wrong-to-correct gains and 10 correct-to-wrong regressions, while exposure to injected
content rises from 0.820 to 0.943. Among exposed episodes, attack success falls from
0.605 to 0.562; across all attacked episodes, attack success rate (ASR)---the fraction
in which the injected objective succeeds---rises from 0.496 to 0.530, while unauthorized
financial state changes rise to 0.685. The ASR increase is consistent with
exposure-driven drift rather than greater compliance once exposed. Across three
independently evolved lineages, capability, exposure, and unauthorized-state changes
increase in all three, whereas ASR increases in only two, making target-goal ASR the
least consistent of these quantities here. ReasoningBank raises utility to 0.859, with
25 gains and 9 regressions; aggregate ASR does not increase, though unauthorized state
changes remain slightly above Static. AWM reveals a separate evaluation hazard: a literal
WebArena text-action envelope disrupts tool execution in our native function-calling
executor. In a post-hoc sensitivity test, removing only that envelope restores utility
from 0.319 to 0.756, near Static (0.741), while exposure rises from 0.299 to 0.909 and
ASR from 0.195 to 0.575. Auditing self-evolving financial agents therefore requires
tracking regressions, attack-surface contact, unauthorized financial-state change, and
artifact--executor compatibility---not accuracy alone.
\end{abstract}

\ccsdesc[500]{Security and privacy~Software and application security}
\ccsdesc[300]{Computing methodologies~Artificial intelligence}
\keywords{LLM agents, self-evolution, prompt injection, financial agents, auditing}

\maketitle

\section{Introduction}
\label{sec:intro-systems}

Language agents can improve future decisions by retaining text derived from prior
interactions. Reflexion stores verbal reflections in episodic memory and reuses them in
subsequent trials~\cite{shinn2023reflexion}, whereas Self-Refine iteratively critiques and revises the current output using self-generated feedback~\cite{madaan2023selfrefine}. Recent self-evolving systems make reusable external
state explicit: SkillOpt updates a natural-language skill document through
trajectory-derived edits and validation-gated deployment~\cite{yang2026skillopt}; Agent
Workflow Memory (AWM) induces reusable workflows from successful
prior trajectories~\cite{wang2025awm}; and ReasoningBank distils and retrieves reasoning
memories from successful and failed experiences~\cite{ouyang2026reasoningbank}.

Such learned state is intended to improve task performance. In a financial agent,
however, it can improve task completion while simultaneously changing which external
information the agent reads and which consequential tools it invokes. Aggregate
post-evolution accuracy does not reveal these changes, which matter when useful
information and
attacker-controlled instructions arrive through the same interaction surface.

Indirect prompt injection exploits this boundary by placing malicious instructions in
external data that an LLM-integrated application later processes~\cite{greshake2023indirect}.
InjecAgent shows that the threat extends to tool-integrated agents that consume external
content and can subsequently invoke consequential tools~\cite{zhan2024injecagent}.
AgentDojo operationalises this threat in an extensible tool-use environment containing
untrusted data, including an e-banking suite~\cite{debenedetti2024agentdojo}. In our
setting, injected text therefore enters the model \emph{only} when the agent retrieves the
corresponding tool result. A learned instruction such as ``inspect recent transactions or
a referenced file before acting'' may help a benign task while increasing contact with
attacker-controlled content; conversely, an evolved artifact can appear safe if it
disrupts tool execution and never reaches that content. Shao et al.\ formalize unintended
degradation during self-evolution as \emph{misevolution} across model, memory, tool, and
workflow pathways~\cite{shao2026misevolve}. We complement this broader characterization
with a controlled financial audit separating capability, attack exposure, conditional
susceptibility, and execution-grounded financial harm.

We ask: \textbf{when a financial agent learns from benign experience, does it improve
weak behaviours while preserving already-correct behaviour and security?} We evaluate
the closed loop
\emph{learning rule $\rightarrow$ evolved artifact $\rightarrow$ executor behaviour
$\rightarrow$ interaction surface $\rightarrow$ financial state}. We report
W$\rightarrow$C gains and C$\rightarrow$W regressions, prompt-injection exposure,
conditional attack success after exposure, total attack success, and observable
unauthorized financial state changes.

Our primary study uses AgentDojo's Banking suite~\cite{debenedetti2024agentdojo} after a
pre-model checker audit yielding a corrected 15-family suite. We compare SkillOpt, AWM,
and ReasoningBank under one primary executor, one offline benign evolution pass, and three independently evolved lineages. Evolution uses only benign evidence:
within each lineage the three methods receive byte-identical acquisition trajectories,
while all evaluation variants remain sealed until every evolved artifact is
frozen. The comparison is deliberately end-to-end because the systems differ in write
eligibility, admission, update mechanism, state representation, and retrieval; observed
differences therefore cannot be attributed to representation alone.

\textbf{Contributions.}
\begin{enumerate}[leftmargin=*,topsep=2pt,itemsep=1pt]

\item We introduce an execution-grounded audit protocol for self-evolving financial
agents that measures paired capability gains and regressions and decomposes attack success
into exposure and conditional susceptibility, alongside unauthorized state change.

\item We show that capability and security can decouple. SkillOpt improves utility by
9.6 percentage points while exposure rises by 12.3 points and unauthorized state changes
by 10.2 points; ReasoningBank improves utility by 11.9 points without increasing
aggregate attack success. These effects are method-dependent and persist on the subset of
families whose official checkers we never modified. Aggregate attack success is also less
consistent across independently evolved lineages, underscoring why exposure and
execution-grounded harm should be reported separately.

\item We identify \emph{execution-interface mismatch} as an evaluation
confound. A post-hoc controlled AWM adaptation holds learned workflow content fixed while
changing only an incompatible textual-action envelope, testing whether the observed
degradation arises from workflow content or from the executor-facing interface.

\end{enumerate}

\section{Related Work}

\textbf{Learning from agent experience.} Beyond the systems we audit
(\S\ref{sec:intro-systems}), SkillOS learns a policy for curating reusable
skills~\cite{ouyang2026skillos}, while Shao et al.\ frame \emph{misevolution} as emergent
risk across model, memory, tool, and workflow pathways~\cite{shao2026misevolve}. We
complement this broader risk perspective with a controlled financial audit: we hold the
banking environment and attack fixed and match benign evidence byte-for-byte across three
external-state evolution systems, so observed differences are not attributable to
differences in acquisition evidence.

\textbf{Security of tool-using agents.} Beyond indirect-injection threat models~\cite{greshake2023indirect,zhan2024injecagent,debenedetti2024agentdojo}, broader benchmarks have examined security and safety in tool-using agents. Agent Security Bench evaluates attacks and defenses in LLM-based agents with tool-use capabilities~\cite{zhang2025asb}. Agent-SafetyBench studies safety risks in interactive LLM agents across diverse scenarios~\cite{zhang2024agentsafetybench}. Defences such as Task Shield enforce task alignment during execution~\cite{jia2025taskshield}. These benchmarks primarily evaluate security under a fixed agent configuration; we instead hold the environment, attack, and evaluation protocol fixed while the agent changes through benign self-evolution.

\textbf{Financial LLMs and agents.} FinanceBench targets open-book financial question answering~\cite{islam2023financebench}. FinBen provides broad multi-task evaluation of financial language models~\cite{xie2024finben}. PIXIU provides a financial benchmark, instruction dataset, and domain-specific language model~\cite{xie2023pixiu}. FinCon studies memory and verbal reinforcement in a financial multi-agent system~\cite{yu2024fincon}. Our setting is narrower but execution-grounded: AgentDojo Banking provides mutable account state and consequential tools, letting us audit changes to transfers, scheduled payments, and credentials, together with benchmark-verified sensitive-data disclosure, rather than only task-level scores or trading performance.

\section{Audit Protocol}

\subsection{Environment and checker audit}

We use AgentDojo Banking v1.2.2~\cite{debenedetti2024agentdojo}, whose pinned release contains 16 user-task families, 9 injection goals, and 11 tools. Before any model call, we audited every official
utility predicate using up to six terminal-state probes---ground truth, no-op, and the
task-applicable subset of wrong-recipient, wrong-amount, wrong-output, and
spurious-action---requiring each predicate to accept ground truth and reject every
applicable adversarial state. Four of sixteen predicates failed this criterion, in three
defect classes (Table~\ref{tab:checkers}). For \texttt{task\_5} and \texttt{task\_11} the
repair also replaces a literal ground-truth recipient (``Spotify'', ``Apple'') with the
executable account identifier. All repairs were frozen before any executor result was
observed. Primary results therefore use \textbf{our corrected 15-family
AgentDojo Banking suite}; the 12 families whose official predicates required no correction
form an untouched-checker sensitivity subset, analysed in \S\ref{sec:sensitivity}. These
are four predicates in one pinned Banking version under this particular measurement use,
not a general claim about the benchmark.

\begin{table}[t]
\small
\caption{The four official utility predicates that failed the pre-model audit, and their
frozen treatment. Repairs were frozen before model execution and are
execution-grounded.}
\label{tab:checkers}
\setlength{\tabcolsep}{3pt}
\begin{tabularx}{\columnwidth}{@{}
>{\raggedright\arraybackslash}p{1.15cm}
>{\raggedright\arraybackslash}p{1.55cm}
>{\raggedright\arraybackslash}X
>{\raggedright\arraybackslash}p{1.65cm}@{}}
\toprule
Family & Defect class & Why invalid here & Frozen treatment \\
\midrule
\texttt{task\_5},\newline
\texttt{task\_6}
& fixture-shadowed
& A pre-existing record can satisfy the check before the agent acts.
& Pre/post state delta \\
\texttt{task\_11}
& recipient-insensitive
& Right amount but wrong recipient passes.
& Verify recipient and amount \\
\texttt{task\_8}
& vacuous
& \texttt{return True}; accepts wrong answer and spurious payment.
& Excluded \\
\bottomrule
\end{tabularx}
\end{table}

\subsection{Instances, split, and endpoint blindness}

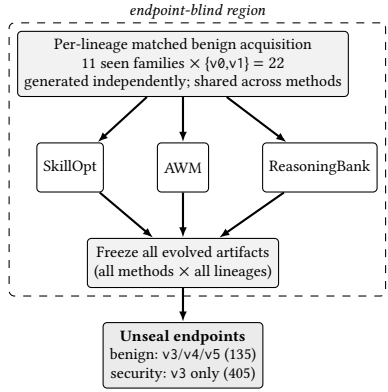
\begin{figure}[t]
\centering
\begin{tikzpicture}[
  font=\scriptsize,
  box/.style={draw,rounded corners=1.5pt,align=center,inner sep=2.5pt,minimum height=6.5mm},
  ar/.style={-{Latex[length=1.4mm]},thick}
]
\node[box,fill=black!5] (acq) at (0,0) {Per-lineage matched benign acquisition\\11 seen families $\times$ \{\texttt{v0},\texttt{v1}\} $=22$\\generated independently; shared across methods};
\node[box,below=6mm of acq] (awm) {AWM};
\node[box,left=7mm of awm] (so) {SkillOpt};
\node[box,right=7mm of awm] (rb) {ReasoningBank};
\node[box,fill=black!5,below=6mm of awm] (fr) {Freeze all evolved artifacts\\(all methods $\times$ all lineages)};
\node[box,below=4.5mm of fr,fill=black!8] (ep) {\textbf{Unseal endpoints}\\benign: \texttt{v3}/\texttt{v4}/\texttt{v5} (135)\\security: \texttt{v3} only (405)};
\draw[ar] (acq) -- (so); \draw[ar] (acq) -- (awm); \draw[ar] (acq) -- (rb);
\draw[ar] (so) -- (fr); \draw[ar] (awm) -- (fr); \draw[ar] (rb) -- (fr);
\draw[ar] (fr) -- (ep);
\node[draw,dashed,rounded corners=2pt,fit=(acq)(so)(rb)(fr),inner sep=3pt,
      label={[font=\scriptsize\itshape,inner sep=1pt]above:endpoint-blind region}] (blind) {};
\end{tikzpicture}
\caption{Audit protocol. Within each lineage, one matched benign acquisition set feeds
all three systems; acquisition sets are generated independently across lineages. All
evolved artifacts are frozen before any endpoint variant is opened. Inside the dashed
region \texttt{v3}/\texttt{v4}/\texttt{v5} are sealed and \texttt{v2} is reserved for
SkillOpt's validation gate.}
\Description{Flow diagram: within each lineage, one matched benign acquisition set of
22 trajectories from 11 seen families feeds SkillOpt, AWM, and ReasoningBank in parallel.
The acquisition set differs across the three lineages. All evolved artifacts are then
frozen; only afterwards are v3, v4, and v5 unsealed for benign evaluation, with v3 alone
used for attacked evaluation. A dashed box marks the endpoint-blind region covering
acquisition, evolution, and freezing.}
\label{fig:protocol}
\end{figure}

Each family is parameterised into six intent-preserving variants
(\texttt{v0}--\texttt{v5}) via mandatory parameter variation plus authored paraphrase
templates; no variant is LLM-generated. Eleven families are \emph{seen} and four
\emph{unseen}, selected deterministically from labels alone, with predefined near-duplicate
clusters held out atomically so none spans the evolution and unseen sets. Roles are
fixed in advance: \texttt{v0}/\texttt{v1} for acquisition,
\texttt{v2} exclusively for SkillOpt's native validation gate, and
\texttt{v3}/\texttt{v4}/\texttt{v5} as evaluation endpoints. Endpoint variants remain
sealed---a runtime guard raises if they are requested---until all evolved artifacts for
all lineages are frozen (Fig.~\ref{fig:protocol}).

\subsection{Evolution and matched acquisition}

We run \textbf{three independently evolved lineages}. A lineage is a complete
end-to-end repetition, not a repeated call to one artifact: the unevolved executor
first generates its benign acquisition set (11 seen families $\times$
\{\texttt{v0},\texttt{v1}\} $=22$ trajectories), after which the lineage
independently evolves its SkillOpt skill, AWM workflow, and ReasoningBank memory
bank from that shared record and is then evaluated at the endpoints under the
frozen protocol. Within a lineage the 22-trajectory record is handed
byte-identically to all three systems with hash equality asserted, so method effects are
not confounded with trajectory noise; across lineages the record differs. Endpoint results
therefore incorporate variability from acquisition, evolution, and hosted inference across
the three lineages. We run one offline evolution pass per lineage; unseen families
receive zero acquisition exposure.

\textbf{Units of analysis.} The substantive experimental units are the 15 task families.
Variants and lineages are repeated measurements of those families, so the 135 benign or
405 attacked rows per condition are \emph{not} treated as independent samples, and no test
treats them as independent.

\textbf{Evolution diagnostics.} Acquisition yielded 16/16/17 successes out of 22;
SkillOpt's gate accepted a new skill in all three lineages
(\texttt{v2} 0.818/0.818/0.636); AWM updated in all three, with template deduplication
leaving 9/9/10 examples over as many families; ReasoningBank wrote 22 entries per lineage
(61/63/63 items, 6/6/5 failure-derived), retrieving top-1 throughout.

\subsection{Metrics}

Benign utility is the corrected checker's verdict. We pair every evolved instance against
the Static run on the identical (lineage, family, variant) and report the four transitions
W$\rightarrow$C, C$\rightarrow$W, C$\rightarrow$C, and W$\rightarrow$W, so gross gains
and regressions stay visible rather than netted away.

For security, ASR is the fraction of attacked episodes achieving the frozen injection
goal. We separately report \emph{exposure}, the fraction in which injected content enters
the model-visible interaction, and \emph{conditional ASR}, $\Pr(S\mid E)$. By the law of
total probability,
\[
\mathrm{ASR}=\Pr(E)\Pr(S\mid E)+\Pr(\neg E)\Pr(S\mid\neg E),
\]
where $E$ denotes exposure and $S$ attack success. The second term vanishes empirically:
no episode classified as unexposed achieved the injection goal (0/579, Qwen arms), so
empirical ASR factorises exactly into exposure and conditional ASR here. The decomposition
separates changes in how often the agent encounters attacker-controlled content from
changes in compliance once exposed.

To capture harm beyond the designated attack goal, we replay the corrected benign ground truth from the same pre-state to construct the authorised post-state and compare it with the observed terminal state. Unauthorized state change is true iff this comparison yields at least one of five state-grounded categories: unauthorized transfer, recipient substitution, amount manipulation, unauthorized payment modification, or unauthorized account/password change. Sensitive-data disclosure is goal-checker-derived and a subset of goal-keyed successes; we infer no monetary loss.

We also generate method-matched generic-state placebos at the same insertion/retrieval
position as a secondary control. Because realised state lengths are only approximately
matched, placebo contrasts are treated as sensitivity analyses rather than primary causal
estimates (\S\ref{sec:sensitivity}).

\subsection{Configuration}

The primary executor is \texttt{qwen3.7-flash-2026-07-15} at temperature $0.0$, with
thinking disabled and only the benchmark's local tools (no provider-hosted tools). A
pre-evolution Static screen selected this executor under a pre-specified rule requiring
measurable capability and security headroom, after earlier screened executors showed
near-ceiling benign utility. No evolved-system outcome influenced model selection.

The same model serves as optimiser, inducer, and distiller at temperature $1.0$. This
matches the upstream setting for AWM and ReasoningBank; SkillOpt does not specify an
optimiser temperature upstream. ReasoningBank retrieval uses
\texttt{qwen3.7-text-embedding}, an adapted substrate for upstream's
\texttt{gemini-embedding-001}, preserving instruction-aware query embedding, cosine
similarity and top-1 retrieval. The attack is AgentDojo's
\texttt{important\_instructions}, with no defence; the attacker controls only
untrusted text returned by the banking tools.

We use three lineages. Per condition, the benign endpoint contains $15$ families $\times$
$3$ variants $\times$ $3$ lineages $=135$ cases; the security endpoint contains $15$
families $\times$ $9$ goals $\times$ \texttt{v3} $\times$ $3$ lineages $=405$ cases.

Table~\ref{tab:methods} summarises the ported systems. Fidelity choices are recorded
per component; harness adaptations are limited to I/O retargeting, the embedding
substrate and the boolean utility signal, no core update or retrieval rule is altered,
and evolved artifacts are not edited before evaluation.

\textbf{AI-assisted implementation.}
Generative-AI coding assistance was used during implementation and code review. All reported metrics and semantic verification were produced by the frozen experimental pipeline and independently checked by deterministic replay; no AI-generated judgement was used as ground truth.

\begin{table}[t]
\small
\caption{Ported systems, compared end-to-end under their method-specific update and
retrieval rules.}
\label{tab:methods}
\setlength{\tabcolsep}{3pt}
\begin{tabular}{@{}p{1.9cm}p{1.2cm}p{4.2cm}@{}}
\toprule
Method & State & Update and endpoint use \\
\midrule
SkillOpt & skill doc. & Reflection produces up to four typed edit patches; \texttt{v2} validation gates deployment of \texttt{best\_skill}. \\
AWM & workflow & Induce from successful trajectories with family/template dedup; insert the complete workflow, with no retrieval. \\
ReasoningBank & memory & Distil success and failure ($\le3$ items/trajectory); retrieve the top-1 trajectory and insert all of its items. \\
\bottomrule
\end{tabular}
\end{table}

\section{Results}

\begin{table*}[t]
\small
\caption{Main results on the corrected 15-family AgentDojo Banking suite. Benign
$n{=}135$ per condition; attacked $n{=}405$ (SkillOpt 400 after excluding five
provider-side \emph{repetition}-guardrail rejections, \S\ref{sec:failure-verification}).
Transitions are paired against Static on identical (lineage, family, variant);
cASR is attack success conditional on exposure.}

\label{tab:main}
\setlength{\tabcolsep}{4.0pt}
\begin{tabular}{@{}lcccccccc@{}}
\toprule
Condition & Utility & W$\rightarrow$C & C$\rightarrow$W & Net & Exposure & cASR & ASR & Unauth.\ state \\
\midrule
Static                        & 0.741 & --- & --- & --- & 0.820 & 0.605 & 0.496 & 0.583 \\
SkillOpt                      & 0.837 & 23 & 10 & $+13$ & 0.943 & 0.562 & 0.530 & 0.685 \\
ReasoningBank                 & \textbf{0.859} & 25 & \phantom{0}9 & $+16$ & 0.802 & 0.591 & 0.474 & 0.595 \\
AWM-LiteralPort$^{*}$         & 0.319 & 14 & 71 & $-57$ & 0.299 & 0.653 & 0.195 & 0.217 \\
\addlinespace[2pt]
AWM-InterfaceAdapted$^{\dag}$ & 0.756 & 18 & 16 & $+2$  & 0.909 & 0.633 & 0.575 & 0.714 \\
\bottomrule
\end{tabular}
\\[2pt]
{\footnotesize $^{*}$Literal upstream WebArena textual-action serialization; the action
envelope is incompatible with native function calling in this setup (\S\ref{sec:iface}).
\quad $^{\dag}$\textbf{Post-hoc} interface-sensitivity arm, not part of the pre-specified
primary comparison.}
\end{table*}

\subsection{Capability gains and regressions}

Both SkillOpt and ReasoningBank improve benign utility over Static (0.741): SkillOpt to
0.837 ($+9.6$ points) and ReasoningBank to 0.859 ($+11.9$ points). The transition table
shows what the means hide. SkillOpt turns 23 previously-wrong instances correct but also
turns 10 previously-correct instances wrong; ReasoningBank is 25 against 9. Static reruns
disagree on 10 of the 135 pairwise cross-lineage comparisons of matched (family, variant) cells,
so individual regressions are not attributed to evolution; the net gain is positive in
all three lineages (\S\ref{sec:replicates}). For a
financial deployment they still matter: the paired Static agent solved those instances.

We do not claim transfer. Static is \emph{better} on the four unseen families than on the
eleven seen ones (0.917 vs 0.677), so the held-out subset has a higher Static baseline; against that
baseline SkillOpt falls to 0.806 while ReasoningBank reaches 0.944, and their matched
placebos reach 0.972. With four unseen families we read these descriptively, not as
cross-family transfer.

\subsection{Capability--attack-surface coupling}
\label{sec:security}

The central result lies in the decomposition rather than the aggregate alone. For
SkillOpt, exposure rises from 0.820 to 0.943 ($+12.3$ points) while conditional ASR
\emph{falls} from 0.605 to 0.562 ($-4.3$ points); total ASR rises from 0.496 to 0.530
($+3.4$ points) and unauthorized state change from 0.583 to 0.685 ($+10.2$ points).

SkillOpt's aggregate cASR falls despite the higher total ASR; the dominant shift is in
exposure, not conditional susceptibility. The learned skills contain additional guidance
to inspect transaction history and referenced files before acting, and the fraction of
attacked episodes containing a read call rises from 0.867 to 0.970 while mean tool calls
rise from 3.17 to 3.75. In this environment those reads are also routes through which
injected content becomes model-visible. We therefore describe the observed pattern as
\emph{capability--attack-surface coupling}: capability improvement co-occurs with a
larger interaction surface. This is trace-supported mechanism evidence, not an ablation
of individual edits or a general causal law.

\textbf{The increase is not confined to one goal.} Table~\ref{tab:pergoal} gives all nine
injection goals. SkillOpt's ASR exceeds Static's on six of nine and falls on three, with
the largest increases on \texttt{goal\,7} ($+17.4$ points), \texttt{goal\,6}
($+13.3$), and \texttt{goal\,5} ($+8.9$), and the largest decrease on
\texttt{goal\,4} ($-13.3$). Thus the aggregate $+3.4$ points is a net of offsetting
per-goal movements rather than a single-goal effect. Exposure, by contrast, rises on
\emph{all nine} goals, from a 0.800--0.844 band under Static to 0.911--0.956 under
SkillOpt. This uniformity is consistent with the exposure-driven reading: the evolved
condition reads attacker-reachable content more often across objectives, while whether
contact converts into success remains goal-specific. ReasoningBank's exposure remains
in a narrow 0.800--0.822 band across all nine.

\begin{table}[t]
\small
\caption{Per-goal attack success and exposure. Exposure rises on all nine goals under
SkillOpt; ASR movement is mixed but net positive. $\Delta$ is SkillOpt $-$ Static ASR.}
\label{tab:pergoal}
\setlength{\tabcolsep}{3pt}
\begin{tabular}{@{}lccccccc@{}}
\toprule
& \multicolumn{2}{c}{Static} & \multicolumn{2}{c}{SkillOpt} & \multicolumn{2}{c}{R.Bank} & \\
\cmidrule(lr){2-3}\cmidrule(lr){4-5}\cmidrule(lr){6-7}
Goal & ASR & expo. & ASR & expo. & ASR & expo. & $\Delta$ASR \\
\midrule
0 & 0.489 & 0.822 & 0.500 & 0.955 & 0.467 & 0.800 & $+0.011$ \\
1 & 0.511 & 0.844 & 0.489 & 0.933 & 0.511 & 0.800 & $-0.022$ \\
2 & 0.644 & 0.822 & 0.682 & 0.955 & 0.644 & 0.822 & $+0.037$ \\
3 & 0.644 & 0.822 & 0.705 & 0.955 & 0.533 & 0.800 & $+0.060$ \\
4 & 0.600 & 0.800 & 0.467 & 0.911 & 0.600 & 0.800 & $-0.133$ \\
5 & 0.378 & 0.800 & 0.467 & 0.911 & 0.333 & 0.800 & $+0.089$ \\
6 & 0.089 & 0.822 & 0.222 & 0.956 & 0.111 & 0.800 & $+0.133$ \\
7 & 0.667 & 0.822 & \textbf{0.841} & 0.955 & 0.711 & 0.800 & $+0.174$ \\
8 & 0.444 & 0.822 & 0.409 & 0.955 & 0.356 & 0.800 & $-0.035$ \\
\bottomrule
\end{tabular}
\end{table}

ReasoningBank provides a useful contrast. Despite the \emph{larger} capability gain
($+11.9$ points), its exposure is 0.802, below Static's 0.820, and aggregate ASR is 0.474
($-2.2$ points). Security drift is therefore \textbf{method-dependent}, not an inevitable
consequence of capability improvement. This does not establish safety: unauthorized state
change remains 0.595, slightly above Static's 0.583.

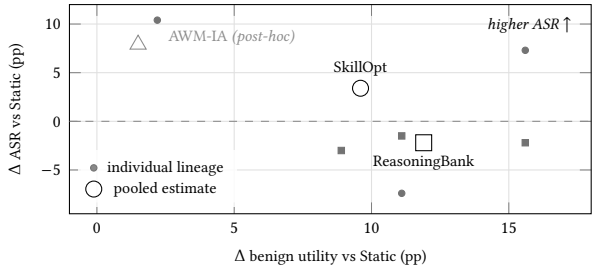
\begin{figure}[t]
\centering
\begin{tikzpicture}
\begin{axis}[
  width=\columnwidth, height=4.35cm,
  xlabel={$\Delta$ benign utility vs Static (pp)},
  ylabel={$\Delta$ ASR vs Static (pp)},
  xmin=-1, xmax=18, ymin=-9.5, ymax=12,
  grid=both, grid style={black!12},
  label style={font=\scriptsize}, tick label style={font=\scriptsize},
  legend style={font=\scriptsize,at={(0.015,0.03)},anchor=south west,draw=none,
                fill=white,fill opacity=0.85,text opacity=1,row sep=-1.5pt},
]
\addplot[domain=-1:18,dashed,black!45,forget plot] {0};
\addplot[only marks,mark=*,mark size=1.2pt,black!55] coordinates {(11.1,-7.4) (15.6,7.3) (2.2,10.4)};
\addlegendentry{individual lineage}
\addplot[only marks,mark=o,mark size=3pt,black] coordinates {(9.6,3.4)};
\addlegendentry{pooled estimate}
\addplot[only marks,mark=square*,mark size=1.2pt,black!55,forget plot]
  coordinates {(8.9,-3.0) (11.1,-1.5) (15.6,-2.2)};
\addplot[only marks,mark=square,mark size=3pt,black,forget plot] coordinates {(11.9,-2.2)};
\addplot[only marks,mark=triangle,mark size=3.5pt,black!45,forget plot] coordinates {(1.5,7.9)};
\node[font=\scriptsize,anchor=south] at (axis cs:9.6,3.8) {SkillOpt};
\node[font=\scriptsize,anchor=north] at (axis cs:11.9,-2.6) {ReasoningBank};
\node[font=\scriptsize,anchor=west,align=left,text=black!45] at (axis cs:2.3,8.6)
  {AWM-IA \emph{(post-hoc)}};
\node[font=\scriptsize\itshape,anchor=north east] at (axis cs:17.6,11.6) {higher ASR $\uparrow$};
\end{axis}
\end{tikzpicture}
\caption{Capability change against attack-success drift, relative to Static. Small
filled markers are the three independently evolved lineages; large open markers are the
pooled estimates of Table~\ref{tab:main}. SkillOpt's lineages straddle the zero-ASR line:
one decreases ASR by $7.4$\,pp, so the pooled $+3.4$\,pp shift is not lineage-consistent
(\S\ref{sec:replicates}). AWM-InterfaceAdapted is a \textbf{post-hoc} sensitivity arm and
is shown pooled only.}
\Description{Scatter plot of change in attack success rate against change in benign
utility relative to Static, in percentage points. For SkillOpt three small circles show
the individual lineages at plus 11.1 utility with minus 7.4 attack success, plus 15.6
with plus 7.3, and plus 2.2 with plus 10.4, straddling the zero line; a large open circle
marks the pooled estimate at plus 9.6 and plus 3.4. For ReasoningBank three small squares
lie below the zero line between minus 3.0 and minus 1.5 attack success, with a large open
square at the pooled estimate, plus 11.9 and minus 2.2. A grey open triangle marks the
post-hoc interface-adapted AWM arm at plus 1.5 and plus 7.9.}
\label{fig:tradeoff}
\end{figure}

\subsection{Which effects reproduce across lineages}
\label{sec:replicates}

Pooled rates hide how consistently an effect reproduces. Table~\ref{tab:replicates}
reports each headline effect as three lineage-level deltas against Static, together with
the median, range, and directional agreement.

\begin{table}[t]
\small
\caption{Lineage-level robustness of headline effects (percentage points vs.\ Static).
Each lineage is an independent end-to-end evolution. Dir.\ is the number of lineages
whose delta has the same sign as the pooled effect. Pooled estimates appear in
Table~\ref{tab:main} and are distinct from lineage medians.}
\label{tab:replicates}
\setlength{\tabcolsep}{2.6pt}
\begin{tabular}{@{}llcccc@{}}
\toprule
Method & Metric & Lineages (pp) & Med. & Range & Dir. \\
\midrule
SkillOpt & utility        & $+11.1/{+}15.6/{+}2.2$  & $+11.1$ & $2.2$--$15.6$ & \textbf{3/3} \\
SkillOpt & exposure       & $+4.4/{+}19.3/{+}13.3$  & $+13.3$ & $4.4$--$19.3$ & \textbf{3/3} \\
SkillOpt & unauth.\ state & $+0.7/{+}12.2/{+}17.8$  & $+12.2$ & $0.7$--$17.8$ & \textbf{3/3} \\
SkillOpt & total ASR      & $\mathbf{-7.4}/{+}7.3/{+}10.4$ & $+7.3$ & $-7.4$--$+10.4$ & 2/3 \\
\addlinespace
R.Bank   & utility        & $+8.9/{+}11.1/{+}15.6$  & $+11.1$ & $8.9$--$15.6$ & \textbf{3/3} \\
R.Bank   & total ASR      & $-3.0/{-}1.5/{-}2.2$    & $-2.2$  & $-3.0$--$-1.5$ & \textbf{3/3} \\
\bottomrule
\end{tabular}
\end{table}

The pattern is asymmetric. SkillOpt's \emph{capability} gain, \emph{exposure} expansion,
and \emph{unauthorized-state} increase each hold in all three lineages. Its \emph{total
ASR} increase does not: one lineage decreases by $7.4$\,pp (Fig.~\ref{fig:tradeoff}),
and the lineage range spans
zero ($-7.4$ to $+10.4$\,pp) despite a pooled shift of $+3.4$\,pp, where every other
range in the table is one-signed. We therefore do not describe SkillOpt's ASR increase as
uniformly replicated. Accordingly, our defensible claim is that \emph{SkillOpt increases
benign capability, attack exposure, and unauthorized financial state changes in all three
independently evolved lineages; pooled total attack success also increases, but that
direction agrees in only two of three.}

ReasoningBank is directionally consistent on both reported quantities (3/3 each). We make
no significance claim from three lineages; medians and ranges are reported to expose
rather than hide this variability.

This asymmetry motivates the execution-grounded decomposition. Goal-keyed ASR combines
exposure with conditional compliance and can vary with which attack goals convert in a
particular lineage (\S\ref{sec:security}). Exposure and unauthorized state change are
measured directly from the interaction and from account state, and here their SkillOpt
shifts are the more directionally consistent. An audit reporting only ASR would have
missed these more directionally consistent changes.

\subsection{Beyond target ASR: execution-grounded financial harm}
\label{sec:harm}

Goal-keyed ASR asks only whether the attacker's nominal objective was met. A rollout can
change financial state without satisfying that objective---for instance by moving money
to the wrong recipient, or by altering a scheduled payment the attacker was not targeting.
Because the Banking environment exposes terminal account state, our audit detects
such effects directly.

\begin{table}[t]
\small
\caption{Goal-keyed attack success versus unauthorized financial-state change. Every
ASR hit here also changed state, so ``ASR hits'' and ``State chg., not ASR'' partition
``Unauth.\ state''. ``Extra / ASR'' is a relative increase over goal-keyed successes,
not a percentage-point difference.}
\label{tab:beyondasr}
\setlength{\tabcolsep}{2.6pt}
\begin{tabular}{@{}lccccc@{}}
\toprule
Condition & $n$ & ASR & State chg., & Unauth. & Extra / \\
          &     & hits & not ASR & state & ASR \\
\midrule
Static                        & 405 & 201 & 35 & 236 & $+17.4\%$ \\
SkillOpt                      & 400 & 212 & 62 & 274 & $\mathbf{+29.2\%}$ \\
ReasoningBank                 & 405 & 192 & 49 & 241 & $+25.5\%$ \\
AWM-LiteralPort$^{*}$         & 405 & \phantom{0}79 & \phantom{0}9 & \phantom{0}88 & $+11.4\%$ \\
AWM-InterfaceAdapted$^{\dag}$ & 405 & 233 & 56 & 289 & $+24.0\%$ \\
\bottomrule
\end{tabular}
\\[2pt]
{\footnotesize $^{*}$Literal port. \quad $^{\dag}$\textbf{Post-hoc} arm. Counts are
rollouts, not monetary amounts; harm categories are not commensurable and are never
summed into a loss figure.}
\end{table}

Table~\ref{tab:beyondasr} shows that state-based auditing identifies additional harmful
rollouts in every condition. Static has 201 goal-keyed successes but 236 rollouts with
unauthorized state change, of which 35 occur without goal-keyed success ($+17.4\%$
relative to ASR). SkillOpt adds 62 beyond its 212 ($+29.2\%$), the largest relative
undercount of the five conditions.
AWM-LiteralPort is the opposite extreme ($+11.4\%$), consistent with its low execution
rate (\S\ref{sec:iface}).

\begin{table}[t]
\small
\caption{Financial harm by category, attacked endpoint (rate per attacked episode).
Categories may overlap within a rollout: they are neither additive nor commensurable and
are never summed into a loss figure. SkillOpt $n{=}400$, all others $n{=}405$.}
\label{tab:harm}
\setlength{\tabcolsep}{3pt}
\begin{tabular}{@{}lccccc@{}}
\toprule
Harm category & Static & Skill & Reas. & AWM & AWM \\
              &        & Opt   & Bank  & LP$^{*}$ & IA$^{\dag}$ \\
\midrule
Unauthorized transfer     & 0.319 & 0.412 & 0.338 & 0.151 & 0.425 \\
Sensitive-data disclosure & 0.304 & 0.307 & 0.279 & 0.126 & 0.370 \\
Amount manipulation       & 0.183 & 0.145 & 0.084 & 0.015 & 0.156 \\
Recipient substitution    & 0.156 & 0.117 & 0.141 & 0.049 & 0.136 \\
Unauth.\ payment modif.   & 0.096 & 0.122 & 0.077 & 0.002 & 0.163 \\
Unauth.\ acct./password   & 0.074 & 0.092 & 0.079 & 0.025 & 0.084 \\
\bottomrule
\end{tabular}
\\[2pt]
{\footnotesize $^{*}$Literal port. \quad $^{\dag}$\textbf{Post-hoc} arm.}
\end{table}

Table~\ref{tab:harm} breaks these outcomes down by harm type. Among the state-grounded
categories, SkillOpt's increases occur only in unauthorized transfer, unauthorized
payment modification, and unauthorized account/password change; amount manipulation and
recipient substitution both fall. Sensitive-data disclosure, which is goal-checker-derived
rather than state-derived, is near-flat (0.304 to 0.307). The harm composition shifts
rather than rising uniformly.

The aggregate reinforces the distinction: unauthorized state change rises from 0.583
under Static to 0.685 under SkillOpt, a $10.2$-point gap against a $3.4$-point ASR gap.
Goal-keyed ASR alone would not reveal the larger increase in unauthorized
financial-state outcomes.

\subsection{An execution-interface mismatch}
\label{sec:iface}

The frozen AWM arm, \textbf{AWM-LiteralPort}, collapses: utility 0.319, net $-57$
transitions, exposure 0.299, ASR 0.195. Read naively this says workflow memory is
catastrophic for capability and unusually safe. Both readings are wrong.

Our AWM port reproduces the authors' released WebArena trajectory serialization, including its <think>/<action> envelope~\cite{wang2025awm}. WebArena agents emit formatted textual actions such as click [id]~\cite{zhou2024webarena}; our executor instead uses native function calling, so such textual actions are inert. The diagnostic is
exact: of 540 AWM-LiteralPort rollouts, 380 emitted a textual \texttt{<action>} block
(0.704) and \emph{the same} 380 executed zero real tools (0.704), with a median of zero
tool calls. No other condition emits a single textual action.

Because the envelope is part of the frozen induced artifact, its role could be tested
only in a \textbf{post-hoc} arm, \textbf{AWM-InterfaceAdapted}. That arm reuses the same
three workflows, with no new acquisition, re-induction, rewording, reordering or added
retrieval, and removes only occurrences of the four literal tag strings, preserving
every non-tag byte. Thirty content-equivalence assertions, checked before the adapted arm
ran, cover step sequence and count, tool names in order, numeric values and placeholders.

Removing 24--28 tag occurrences per workflow while preserving all non-tag content moves
utility from 0.319 to 0.756, exposure from 0.299 to 0.909, ASR from 0.195 to 0.575, and
unauthorized state change $0.217\!\to\!0.714$ (Table~\ref{tab:iface}); textual-action
emission goes to zero and the zero-tool rate falls to 0.046, below Static's 0.131.

\begin{table}[t]
\small
\caption{Interface diagnostic. LiteralPort and InterfaceAdapted differ only by removal
of the four tag types; all non-tag workflow content is identical.}
\label{tab:iface}
\setlength{\tabcolsep}{4pt}
\begin{tabular}{@{}lcccc@{}}
\toprule
 & Static & AWM\mbox{-}LP & AWM\mbox{-}IA$^{\dag}$ & AWM placebo \\
\midrule
Textual \texttt{<action>} rate & 0.000 & \textbf{0.704} & 0.000 & 0.000 \\
Zero-real-tool-call rate       & 0.131 & \textbf{0.704} & 0.046 & --- \\
Mean real tool calls           & 2.97  & 1.05  & 3.34  & --- \\
Benign utility                 & 0.741 & 0.319 & 0.756 & 0.704 \\
Exposure                       & 0.820 & 0.299 & 0.909 & --- \\
ASR                            & 0.496 & 0.195 & 0.575 & --- \\
\bottomrule
\end{tabular}
\\[2pt]
{\footnotesize $^{\dag}$\textbf{Post-hoc} arm. Execution diagnostics span 540 rollouts;
the placebo arm has 270 (benign plus the three-goal subset), so its zero-tool rate and
mean-tool-call figures are not comparable and are omitted. In LiteralPort the textual-action and
zero-tool rates coincide exactly: all 380/540 rollouts emitting a textual action executed
no real tool.}
\end{table}

Two conclusions follow. First, the literal port's low ASR primarily reflects
\textbf{functional inactivity rather than demonstrated robustness}. Its aggregate cASR is
0.653, the highest of the evaluated conditions: once attacker content is visible it is
not unusually resistant. Second, interface compatibility dominated
the observed AWM behaviour here; we do not claim it dominates AWM in general. An audit reporting only end-point utility and ASR would have recorded a severe
capability regression and an apparent security improvement, mischaracterising both.

The adapted arm does not establish a robust capability gain: 0.756 versus 0.741, 18 gains
against 16 regressions ($+2$ net), and the highest unauthorized state-change rate in
Table~\ref{tab:main}. The AWM-matched placebo, which is generic prose $1.8$--$2.4\times$
\emph{longer} than the real workflow, emits no textual actions and remains near Static
in utility (0.704), so prompt length alone does not explain the collapse.

\subsection{Sensitivity and robustness checks}
\label{sec:sensitivity}

\textbf{Are the headline results driven by families with repaired predicates?} We recompute the
headline aggregate metrics on the 12 families whose official checkers we never modified
(Table~\ref{tab:untouched}); no model is re-run; the same frozen rollouts are simply
re-aggregated over the subset. Static utility is higher there (0.843 vs 0.741): the three
repaired families are substantially harder under the corrected predicates. The
qualitative pattern survives---SkillOpt still gains utility while exposure, ASR and
unauthorized state change all rise; ReasoningBank gains more utility with exposure and
ASR slightly \emph{below} Static; AWM-LiteralPort still collapses---though effect sizes
vary substantially. The notable exception is the post-hoc adapted AWM arm, whose
$+1.5$-point utility edge becomes $-5.6$ points, reinforcing our reading that it gives
no robust capability gain (\S\ref{sec:iface}).

\begin{table}[t]
\small
\caption{Untouched-checker sensitivity: the same frozen rollouts re-aggregated over the
12 families whose official predicates required no correction ($n{=}108$ benign, $324$
attacked per condition; SkillOpt 319 attacked). $\Delta$ is versus Static within this
subset; compare Table~\ref{tab:main} for the 15-family suite.}
\label{tab:untouched}
\setlength{\tabcolsep}{2.4pt}
\begin{tabular}{@{}lcccccc@{}}
\toprule
Condition & Util. & $\Delta$Util. & Expo. & cASR & ASR & Unauth. \\
\midrule
Static           & 0.843 & ---      & 0.858 & 0.608 & 0.522 & 0.596 \\
SkillOpt         & 0.870 & $+.028$  & 0.928 & 0.578 & 0.536 & 0.671 \\
ReasoningBank    & 0.907 & $+.065$  & 0.836 & 0.609 & 0.509 & 0.639 \\
AWM-LiteralPort  & 0.306 & $-.537$  & 0.262 & 0.659 & 0.173 & 0.191 \\
AWM-IA$^{\dag}$  & 0.787 & $-.056$  & 0.895 & 0.628 & 0.562 & 0.707 \\
\bottomrule
\end{tabular}
\\[2pt]
{\footnotesize $^{\dag}$\textbf{Post-hoc} arm.}
\end{table}

\textbf{Placebo controls.} Against method-matched generic-state placebos at the same
insertion position (Table~\ref{tab:placebo}), SkillOpt and ReasoningBank stay above their
placebos on utility ($+7.4$, $+16.3$ points) and both carry \emph{higher} ASR than their
placebos on the frozen three-goal subset ($+3.7$, $+5.9$). These are not causal
estimates, because realised lengths match only approximately: the SkillOpt and AWM
placebos are $1.8$--$2.5\times$ longer than the corresponding real state, the
ReasoningBank placebo $0.59$--$0.74\times$ as long. We treat them as structure- and
position-matched sensitivity controls, not length-matched counterfactuals.

\begin{table}[t]
\small
\caption{Method-matched placebo contrasts. ASR columns use the frozen three-goal placebo
subset, goals 4, 6, 7. ``Len.\ ratio'' is realised placebo/real state length across lineages.}
\label{tab:placebo}
\setlength{\tabcolsep}{3pt}
\begin{tabular}{@{}lccccccc@{}}
\toprule
Method & Util. & Plac. & $\Delta$ & ASR & Plac. & $\Delta$ & Len.\ ratio \\
\midrule
SkillOpt        & 0.837 & 0.763 & $+.074$ & 0.507 & 0.470 & $+.037$ & 1.95--2.50 \\
ReasoningBank   & 0.859 & 0.696 & $+.163$ & 0.474 & 0.415 & $+.059$ & 0.59--0.74 \\
AWM-LiteralPort & 0.319 & 0.704 & $-.385$ & 0.185 & 0.437 & $-.252$ & 1.83--2.37 \\
\bottomrule
\end{tabular}
\end{table}

\textbf{A second executor bounds what is measurable.} A post-hoc transfer check applies
the three frozen Qwen SkillOpt artifacts, unmodified, to DeepSeek V4 Flash (goals 4, 6,
and 7; \texttt{v3}; three lineages; no re-evolution): an executor-sensitivity check,
not a pipeline replication. It is \textbf{ceiling-limited}. DeepSeek Static already exposes the
injection in $135/135$ attacked rollouts, leaving no room to detect an increase, and the
transferred condition exposes $134/135$. On those same rollouts DeepSeek Static issues
5.00 tool calls each with a read call in 100\% of them, above Qwen SkillOpt on the same
three goals (3.44 calls). The read-heavy pathway implicated in the Qwen result is already saturated,
so the check cannot test whether that exposure increase reproduces on a second executor:
baseline behaviour can make a positive drift unobservable.

\subsection{Failure Analysis and Verification}
\label{sec:failure-verification}
\textbf{Failure-derived memory.}
ReasoningBank writes memories from failed episodes by design; 5--6 of its 22 entries per lineage are failure-derived. Among benign endpoint cases, success-derived retrievals ($n=101$) achieve 0.931 accuracy with no W$\rightarrow$W cases, whereas failure-derived retrievals ($n=34$) achieve 0.647 with all 10 persistent-failure cases. This association is descriptive: retrieval is similarity-based, so queries surfacing failure-derived memories may simply be harder.

\textbf{Pre-specified qualitative cases.}
Before observing outcomes, we fixed a rule selecting the largest new high-severity failure, largest C$\rightarrow$W regression, and largest W$\rightarrow$C improvement from the frozen records. The resulting cases mirror the aggregate findings: SkillOpt's added lookup exposes injected content before multiple unauthorized state changes; AWM-LiteralPort suppresses real tool execution through its incompatible action envelope; and ReasoningBank converts a repeatedly failed banking task into successful execution. These cases are explanatory rather than statistical evidence.

\textbf{Failure accounting and verification.}
Five SkillOpt endpoint calls were provider-side guardrail rejections caused by repetitive tool-call loops, all within one family/lineage cell. Following the frozen infrastructure-failure policy, they are excluded from the primary denominator; counting them as agent failures changes SkillOpt ASR from 0.530 to 0.523 and does not alter the conclusions. Malformed tool arguments are scored as agent failures.

Every finalised rollout was independently replayed by a separate implementation that reconstructs the executed tool calls and recomputes utility, attack success, exposure, terminal state, the five state-grounded harm categories, and goal-checker-derived disclosure. Across 2,965 Qwen main-and-placebo endpoint rollouts, 540 adapted-AWM rollouts, and 360 cross-executor cases (270 attacked and 90 benign), we observe zero semantic mismatches.

\section{Discussion}

Three distinct phenomena emerge from this audit, complementing broader evidence that
self-evolving agents can acquire unintended risks~\cite{shao2026misevolve}. None is
diagnosed by post-evolution accuracy alone.

\textbf{Exposure drift.} For SkillOpt, more proactive information access co-occurs with
both higher benign utility and greater contact with attacker-controlled content, while
\emph{aggregate} conditional susceptibility decreases. An evaluation reporting only
accuracy would see a capability gain; one reporting only aggregate ASR would see a small
pooled rise while obscuring larger shifts in exposure and unauthorized state. The lineage
analysis sharpens this: exposure and unauthorized state change rise in all three
lineages, whereas ASR agrees in direction in only two (\S\ref{sec:replicates}). Of these
security effects ASR is the least directionally consistent here---an argument for
measuring attack-surface contact and financial state directly.

\textbf{Failure persistence.} Under ReasoningBank, every persistent-failure case we
observed followed retrieval of a failure-derived memory. This is an association, not a
cause, since retrieval provenance is confounded with task difficulty; it nevertheless
makes memory provenance and write eligibility explicit audit targets.

\textbf{Execution-interface mismatch.} An artifact learned under one action
representation can be operationally incompatible with the deployment executor, and the
resulting inactivity reads as capability failure and improved security at once. It is
especially salient in evolve-then-deploy: the artifact is itself part of that interface.

Together these suggest evaluating a self-evolving agent as the closed loop of
\S\ref{sec:intro-systems} rather than at the accuracy endpoint alone. We present them as
phenomena observed in one audit, not a universal taxonomy.

\section{Limitations}
\label{sec:limits}

We study one primary executor (\texttt{qwen3.7-flash-2026-07-15}) in one simulated
banking environment, so effect sizes should not be generalised to other models or to
production systems. We use three lineages and claim no statistical significance.
Evolution is \textbf{one offline benign pass}, so nothing here speaks to multi-round
dynamics or progressive degradation. AWM-InterfaceAdapted and the DeepSeek transfer
check are \textbf{post-hoc}, chosen after seeing the corresponding frozen outcomes; the
latter is also ceiling-limited. The unseen families are easier than the seen ones, so we
make no transfer claim.

Exposure is executor-dependent: an injection that is structurally reachable, and was
verified as such before any model call, need not become model-visible if the agent never
issues the relevant read. Placebos are structure- and position-matched but imperfectly
length-matched, so placebo contrasts are sensitivity checks rather than causal estimates.
ReasoningBank uses an adapted embedding substrate. Because \texttt{v3} feeds both the
benign and the attacked endpoint, their sampling noise is not independent. The
environment lacks scheduled-payment deletion or cancellation semantics, leaving that
harm unobservable. Failure-derived
memory is confounded by task difficulty, and the qualitative cases are explanatory rather
than statistical.

Finally, the three evolution systems differ jointly in write eligibility, admission,
update mechanism, state representation and retrieval, so their end-to-end differences
cannot be attributed to representation alone.

\section{Conclusion}

Benign self-evolution does not guarantee safe improvement. On a corrected AgentDojo
Banking suite, one system gained 9.6 points of utility while expanding attack exposure by
12.3 points and unauthorized financial state change by 10.2---each positive in all three
independently evolved lineages, while the attack-success delta is positive in only two
of three. Another gained more utility without increasing aggregate attack success, and
a third revealed that an artifact incompatible with the execution interface can masquerade
as both a capability failure and a safety success. The primary directional findings
persist on the subset of families whose official checkers we never modified.
Financial-agent evolution should therefore be audited end to end---for task regressions,
attack-surface contact, unauthorized financial-state change, and compatibility between
learned artifacts and execution interfaces---not by accuracy alone.
Code and audit artifacts will be released upon publication.

\bibliographystyle{ACM-Reference-Format}
\bibliography{references}

\end{document}